\documentclass[twoside]{article}

\usepackage[accepted]{aistats2025}
\usepackage[dvipsnames]{xcolor}
\usepackage{graphicx}
\usepackage{booktabs}
\usepackage{siunitx}
\usepackage{tabularx}
\usepackage{natbib}
\usepackage{subcaption}
\usepackage{tcolorbox}
\usepackage{url}
\usepackage{enumitem}
\usepackage{hyperref}

\newcommand\tdots{\hbox to 0.9em{.\hss.\hss.}}

\definecolor{britishracinggreen}{rgb}{0.0, 0.26, 0.15}

\begin{document}

\runningtitle{InnerThoughts: Disentangling Representations and Predictions in LLMs}
\runningauthor{Chételat, Cotnareanu, Thompson, Zhang and Coates}

\twocolumn[

\aistatstitle{InnerThoughts: Disentangling Representations\\ and Predictions in Large Language Models}

\aistatsauthor{ Didier Chételat \And Joseph Cotnareanu \And  Rylee Thompson }

\aistatsaddress{ Huawei Noah's Ark Lab \And  McGill University \And Huawei Noah's Ark Lab } 

\aistatsauthor{ Yingxue Zhang \And Mark Coates }

\aistatsaddress{ Huawei Noah's Ark Lab \And  McGill University} 

]

\begin{abstract}
Large language models (LLMs) contain substantial factual knowledge which is commonly elicited by multiple-choice question-answering prompts. Internally, such models process the prompt through multiple transformer layers, building varying representations of the problem within its hidden states. Ultimately, however, only the hidden state corresponding to the final layer and token position are used to predict the answer label. In this work, we propose instead to learn a small separate neural network predictor module on a collection of training questions, that take the hidden states from all the layers at the last temporal position as input and outputs predictions. In effect, such a framework disentangles the representational abilities of LLMs from their predictive abilities. On a collection of hard benchmarks, our method achieves considerable improvements in performance, sometimes comparable to supervised fine-tuning procedures, but at a fraction of the computational cost.
\end{abstract}

\section{Introduction}

\begin{figure*}[t]
    \centering
    \begin{subfigure}[b]{0.5\textwidth}
    \centering
    \includegraphics[height=6.5cm]{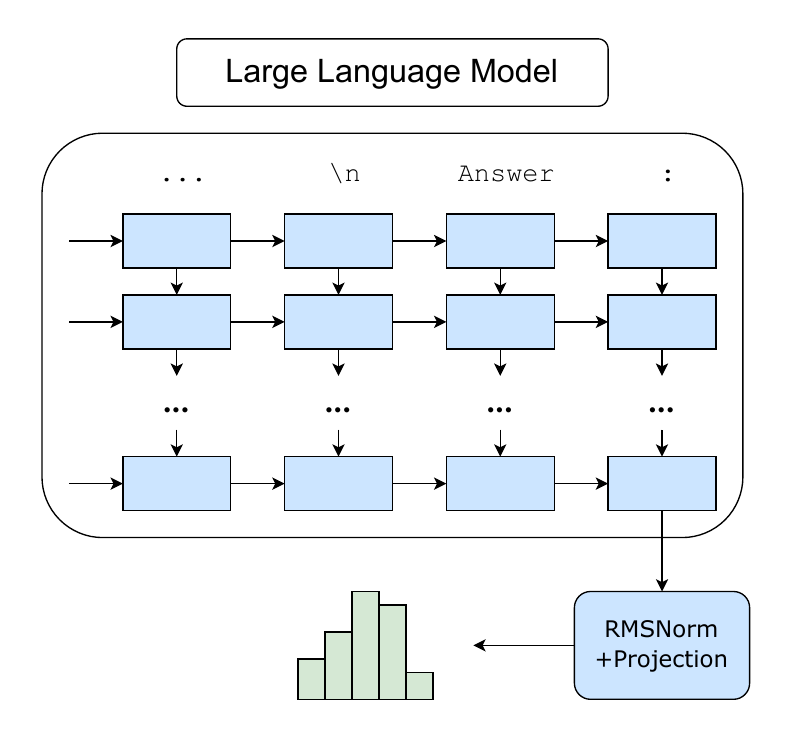}
    \caption{How LLMs (e.g., Llama3) currently operate.}
    \label{fig:method-vanilla}
    \end{subfigure}%
    ~ 
    \begin{subfigure}[b]{0.5\textwidth}
    \centering
    \includegraphics[height=6.5cm]{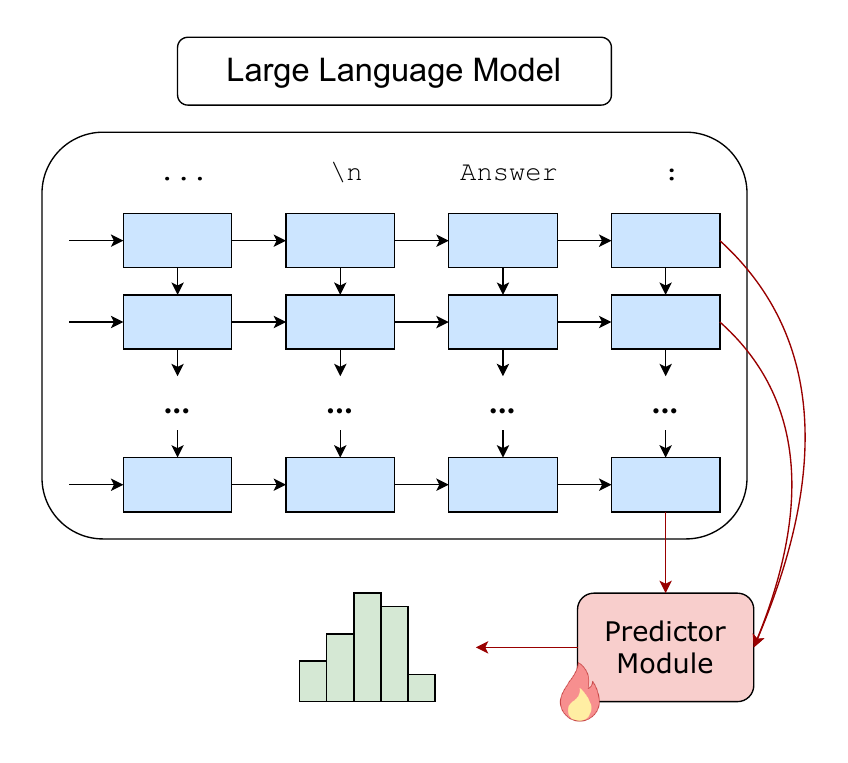}
    \caption{The proposed method.}
    \label{fig:method-innerthoughts}
    \end{subfigure}
    \caption{A comparison of the proposed method with how LLMs currently answer multiple-choice questions. (Left) The LLM is provided with the question and the labeled options, and the probabilities of the labels (`~A', `~B', ...) are computed. These are interpreted as the probabilities of the answer. The probabilities themselves are obtained by taking the hidden state at the last layer and position, passing it through a normalization layer followed by a projection matrix, and taking a softmax. (Right) In our proposed method, the hidden states of all the layers in the last position are sent to a trained predictor module, which directly makes predictions.}
\end{figure*}

\begin{table}[t]
\caption{A multiple-choice question example from the MedMCQA benchmark.}
\label{tab:example}
\vspace{-5pt}
\begin{tcolorbox}[colback=blue!5!white]
\textbf{Question:} Which one of the following statements is true of brain tumours in childhood?\\
A. Is a rare from of malignancy\\
B. Most tumours are below the tentorium\\
C. Hemiparesis is a frequent from of presentation\\
D. Papilloedema is infrequent\\
\textbf{Answer: }B
\end{tcolorbox}
\end{table}

Large Language Models (LLMs) have been found to encode a significant amount of common sense and technical knowledge, which can be elicited by natural language question-answer prompting \citep{petroni2019language, talmor2020olmpics, jiang2021can, carlini2021extracting, ji2023survey}. However, large language models have also been found to suffer from numerous biases when answering questions \citep{ko2020look, jiang2021can, mao2021eliciting, zheng2023large}, which has motivated research into techniques to correct the models to better align their answers with ground truth \citep{zhao2021calibrate, jiang2021can}. The most successful of these methods assume a labeled training set is available, which turns the problem into a form of supervised learning.

In this work, we focus on multiple-choice question answering for simplicity, although the approach taken in this paper could be extended to other formats such as numerical or sequential answers. In a multiple-choice question, the possible answers are finite and provided to the large language model before querying, which means that the language model only has to predict a single token, the correct answer label. An example of such a question is given in Table \ref{tab:example}. Even in this simple format, numerous biases already arise from the order of the choices \citep{pezeshkpour2023large}, the answer labels \citep{zheng2023large}, and the prompt format \citep{robinson2023leveraging}.

The most natural approach to improve performance is to fine-tune the parameters of the language model to better predict the correct label. However, even with parameter-efficient strategies such as LoRA \citep{hu2022lora} and QLoRA \citep{dettmers2024qlora}, this approach remains computationally prohibitive for all but the most sophisticated hardware. Instead, a more accessible alternative has recently been proposed in the form of Linear Probe Calibration \citep{abbas2024enhancing}. Whereas in the standard approach, the logits $q$ of the possible answer labels are extracted from the language model and transformed by a softmax activation to produce the answer distribution $p=\text{softmax}(q)$, Linear Probe Calibration instead proposes to learn a linear transformation $Aq+b$ of the logits on the training set, to correct them before the softmax layer, so that $p_\text{LinProbe}=\text{softmax}(Aq+b)$. Although not originally presented as such, this approach is really equivalent to training a logistic regression model on the logits. 

Now, these logits themselves are obtained, in current state-of-the-art LLMs, from a two-step process. First, the question prompt is processed through multiple transformer layers, each time producing a sequence of hidden states. Second, the hidden state corresponding to the final layer and token position is processed by a normalization layer followed by an embedding layer, to produce the logits for the next token, which is interpreted as the answer. This process suggests that richer information could be obtained from the inner thoughts of the model, rather than the logits it outputs.

In this work, we propose to attach to the LLM a module that makes predictions about the answer label from the hidden states produced by every layer at the final token position. This predictor network is trained on a collection of labeled examples, while keeping the parameters of the language model frozen. Because such a predictor appears after the transformer blocks, training requires no backward passes through the LLM. Only a single forward pass is required to initially extract the hidden states, keeping the training costs to a minimum. 
In a sense, our decomposition into two models seen as a strategy to disentangle the abilities of pretrained LLMs in creating general representations, from its abilities to complete tasks using these representations. We keep the former, but we replace the latter by a small, task-specific module.
On a collection of hard benchmarks, our approach, which we call \textbf{InnerThoughts}, is shown to provide substantial gains beyond previous calibration approaches, at a fraction of the cost of parameter-efficient fine-tuning methods like QLoRA \citep{dettmers2024qlora}.

We can summarize our contributions as follows.
\begin{itemize}
    \item We propose a novel approach to improve large language model accuracy on multiple-choice question-answering tasks, where a small feedforward neural network is trained on the hidden states from all layers at the last token position to predict answers. \\
    \item We show that this approach improves substantially over previous calibration approaches, approaching parameter-efficient fine-tuning in some cases, but at a fraction of the computational cost.
    \item In a subsequent analysis, we identify that the benchmarks where the biggest gains are obtained are precisely those where the language model's classification confidence in its answers is the lowest. We also identify contributions from specific layers that justify our approach.
\end{itemize}

\section{Related Work}

The closest work to ours is Linear Probe Calibration \citep{abbas2024enhancing}, which proposes to learn, using a training set, a logistic regression model that maps logits produced by large language models to multiple-choice answers. This work is related to the broader literature on reducing the numerous biases present with similar tasks, which is called ``calibration'' (not to be confused with statistical calibration).
Contextual Calibration \citep{zhao2021calibrate} attempts to correct for biases in the template by calculating logits for questions with dummy choices (such as [N/A] or the empty string) -- these calculated logits are then subtracted from real question logits during inference.
Whereas Contextual Calibration attempts to correct for bias towards labels, Domain-Context Calibration \citep{fei2023mitigating} attempts to correct for the influence of the distribution of words in the question vocabulary. The method involves computing the logits of meaningless questions and choices created from random strings from the question and choice vocabulary. As in Contextual Calibration, the calculated logits are subtracted from the real question logits during inference. 
Prototypical Calibration \citep{han2023prototypical} proposes that instead of taking the argmax of the output probabilities as the decision, the output probabilities should be clustered by a Gaussian Mixture Model and the label should be selected as the most common label in the cluster.
Finally, Batch Calibration \citep{zhou2023batch} proposes to average the output distributions over the test set, which they subtract from real question logits at test time. This can be done, for example, in a streaming setting, where test examples are progressively observed.

\begin{figure*}[t]
    \centering
    \includegraphics[width=0.8\textwidth]{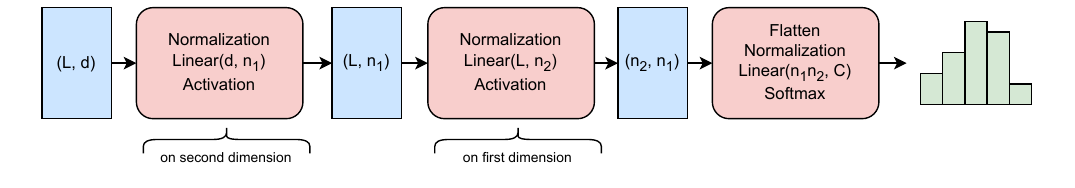}
    \caption{A diagram of our suggested predictor module architecture. The network takes inputs of the shape $(L, d)$, where $L$ is the number of layers in the large language model and $d$ is the dimension of the hidden states, and outputs a distribution over $C$ choices.}
    \label{fig:predictor-diagram}
\end{figure*}

These calibration methods do not require labeled training sets, or only small ones, so that they can be best themed unsupervised or semi-supervised. In contrast, by their usage of larger training sets for learning new modules, our method and Linear Probe Calibration are more closely related to parameter-efficient fine-tuning methods using adapters, which add a few trainable parameters within each transformer block of the LLM.
The most popular variants of this approach are LoRA \citep{hu2022lora} and QLoRA \cite{dettmers2024qlora}, which propose to learn low-rank additive complements to the model parameters. Other popular variants include ``bottleneck'' adapters \citep{houlsby2019parameter, pfeiffer2020mad, he2021towards}, which insert small feedforward layers within the attention blocks; Compacter, \citep{karimi2021compacter} which add linear layers with weights decomposable into sums of Kronecker products of small matrices; (IA)$^3$ \citep{liu2022few}, which multiplies parameters by adapter matrices; and the more recent ReFT \citep{wu2024reft} approach, which adds linear transformations to the parameters and is inspired by causal interventions in interpretability research \citep{NEURIPS2021_4f5c422f}.
In common to all these methods is that adapter modules are inserted within the layers of the model, which means that new forward and backward passes must be repeatedly taken as the adapters are trained. In contrast, our method only requires a single forward pass through the LLM during the entire training process, which makes it much cheaper and more stable.

There also exist other forms of parameter-efficient fine-tuning that rely on modifying inputs, often with trainable modules. For example, Prompt Tuning \citep{lester2021power} and Prefix Tuning \citep{li2021prefix} propose to tune inputs in embedding space using continuous optimization methods, which can represent information more compactly than tokens. Other variants include using mixtures of these learned embeddings \citep{qin2021learning}, training a module that transforms token embeddings \citep{an2022input}, and tuning prompts with reinforcement learning \citep{deng2022rlprompt}. Again, just like adapter methods, such methods require multiple forward and backward passes and are considerably more expensive and numerically unstable than our technique.

Some prior works have investigated using information in the inner layers for other purposes than question answering. SAPLMA \citep{azaria2023internal} propose to train a small feedforward neural network on an inner layer of an LLM, to predict whether a statement just uttered is likely to be true or false. Early-exit strategies such as Logit Lens \citep{logitlens}, Tuned Lens \citep{belrose2023eliciting} and CALM \citep{schuster2022confident} propose to reduce inference cost by skipping deeper layers when they are not necessary. Linear Shortcut \citep{din2024jump} propose a similar strategy with linear transformations, and also propose an ``alternating mode'' where some inner transformer layers are replaced by linear layers. Finally, DoLa \citep{chuang2023dola} propose to renormalize the token distribution by an inner layer's predicted distribution to improve truthfulness in question-answering tasks.

\section{Methodology}\label{sec:methodology}

We now describe in detail our proposed approach. Our exposition assumes that the LLM follows a decoder-only architecture, which forms the basis of all current state-of-the-art models \citep{minaee2024large}. To make a prediction for a given question, we first provide to the LLM a standardized prompt, which follows the recommendations of~\citet{robinson2023leveraging}. Each prompt is composed of the question prefixed by ``Question:'', followed by the possible answers with alphabetical labels. Some datasets include additional contextual information, in which case it is provided before the question with the prefix ``Passage:''. The prompt is then followed by an answer-triggering sentence, in our case simply ``Answer:''. Examples of prompts for each dataset are provided in Appendix~\ref{sec:example-prompts}.

The LLM processes this prompt as follows. First, the prompt is broken into $K$ tokens $T_1,\tdots,T_K$, which are mapped to vectors $h^{0}_1,\tdots,h^{0}_K$ in a high-dimensional embedding space. This sequence of vectors is then processed by multiple transformer blocks, each block outputting a new sequence of hidden states $h^{l}_1,\tdots,h^{l}_K$, for $l=1,\tdots,L$. The standard approach would then use the hidden state from the last layer at the last position, $h^{L}_{K}$, to predict the logits of the possible answer labels (`~A', `~B', $\tdots$), via a combination of a normalization layer and multiplication by an embedding matrix. The normalization layer varies from model to model, such as LayerNorm for the GPT family \citep{radford2019language, brown2020language} and RMSNorm for the Llama family \citep{touvron2023llama, dubey2024llama}. The final prediction probabilities are obtained by softmaxing the logits. A diagram is provided in Figure \ref{fig:method-vanilla}.

Instead, we propose to replace this final normalization and embedding by a predictor module that takes \emph{all} the hidden states at the last position, ($h^{1}_{K}, \tdots, h^{L}_{K}$), and outputs a probability distribution over the choices. Figure~\ref{fig:method-innerthoughts} illustrates the approach. Although, in principle, any classifier could be used for the purpose, in our experiments we use a small neural network with an architecture similar to MLPMixer~\citep{tolstikhin2021mlp}. This architecture was chosen as it avoids introducing too many parameters compared to alternatives, such as flattening the $L\times d$ vector and using a standard FNN architecture.

\begin{table}[t]
\caption{Number of examples in each dataset.}
\label{tab:dataset-sizes}
\centering
\begin{tabular}{
    l
    >{\raggedleft}p{0.15\columnwidth}
    >{\raggedleft}p{0.20\columnwidth}
    >{\raggedleft\arraybackslash}p{0.17\columnwidth}
    }
\toprule
& \text{Train}
& \text{Validation}
& \text{Test}
\\
\midrule
\textsc{AQuA} & 68226 & 14620 & 14621 \\
\textsc{CosmosQA} & 20209 & 5053 & 2985 \\
\textsc{MMLU} & 99842 & 1531 & 14042 \\
\textsc{MedMCQA} & 127975 & 27423 & 27424 \\
\textsc{LogiQA} & 12567 & 1569 & 1572 \\
\textsc{HellaSwag} & 31924 & 7981 & 10042 \\
\bottomrule
\end{tabular}
\end{table}

In detail, our predictor model takes as inputs tensors of the shape $(L, d)$, where $L$ is the number of layers of the LLM and $d$ is the dimension of the hidden states. The model first applies a block composed of a normalization layer such as LayerNorm or RMSNorm, a linear layer and an activation function such as a ReLU or Swish to reduce the last dimension down to $n_1$ dimensions. Then a second block composed of a normalization layer, a linear layer and an activation function reduces the initial dimension down to $n_2$. The two dimensions are then flattened into a single $n_1\times n_2$ dimensional vector, and a final block composed of a normalization layer, a linear layer and a softmax activation maps this vector to a distribution over the $C$ possible answers. Figure~\ref{fig:predictor-diagram} depicts the architecture and calculation.

We note in passing that this architecture is strictly more expressive than the original LLM model, in the sense that we can recover the original LLM's predictor module with specific parameter and hyperparameter choices. This can be achieved, for example, by setting $n_1=d$ and the first linear layer be an identity matrix with no normalization and activation, $n_2=1$ and the second linear layer be multiplication by the vector $(0, \dots, 0, 1)$ with no normalization and activation, and the final Linear$(d\times1, C)$ layer be the original embedding projection matrix with the same normalization layer (LayerNorm or RMSNorm) as in the large language model.


This predictor module is trained on the training set while keeping the parameters of the LLM fixed. Since it is attached after the transformer blocks, the approach is likely to be more limited in its predictive capacity than adapter-based fine-tuning methods like LoRA \citep{hu2022lora}. On the other hand, unlike those methods, only a single forward pass and no expensive backward passes through the LLM are required for training, making them significantly cheaper. 

In general, one could imagine having at disposal a collection of pretrained predictor modules, one for each task at hand, with the original normalization and embedding block serving as a general-purpose ``language modeling'' predictor. This approach would tie in with analogous visions for adapter-based methods such as AdapterHub \citep{pfeiffer2020adapterhub}.

\begin{table*}[t]
\caption{Accuracy of the competing methods on the benchmarks, using Llama3 70B \citep{dubey2024llama} as the LLM. We also report the width of bootstrapped 95\% confidence intervals. For every best entry, we bold the number and perform a one-sided Wilcoxon signed-rank test to determine whether the difference in results with the Direct method is significantly greater than 0. An asterisk (*) next to a bolded number denotes a p-value smaller than 0.05. Finally, we report the accuracy of the model fine-tuned with QLoRA \citep{dettmers2024qlora}, which acts as an upper-bound on the performance.}
\vspace{1mm}
\label{tab:results-raw}
\centering
\aboverulesep = 0.5mm
\belowrulesep = 1mm
\setlength{\tabcolsep}{4.5pt} 
\sisetup{detect-weight=true}
\begin{tabularx}{\textwidth}{
    l
    S[table-format=3.2]@{\:\( \pm \)\,} S[table-format=1.1]
    S[table-format=3.2]@{\:\( \pm \)} S[table-format=1.1]
    S[table-format=2.2]@{\:\( \pm \)\,} S[table-format=1.1]
    S[table-format=3.2]@{\:\( \pm \)} S[table-format=1.1]
    S[table-format=3.2]@{\:\( \pm \)\,} S[table-format=1.1]
    S[table-format=3.2]@{\:\( \pm \)} S[table-format=1.1]
    }
\toprule
& \multicolumn{2}{c}{\textsc{AQuA}}
& \multicolumn{2}{c}{\textsc{CosmosQA}}
& \multicolumn{2}{c}{\textsc{MMLU}}
& \multicolumn{2}{c}{\textsc{MedMCQA}}
& \multicolumn{2}{c}{\textsc{LogiQA}}
& \multicolumn{2}{c}{\textsc{HellaSwag}}
\\
\midrule
\text{Direct}
& 36.47  & 0.8   
& 78.38  & 1.4   
& 73.88  & 0.8   
& 83.59  & 0.4   
& 64.25  & 2.4   
& 82.20  & 0.7   
\\
\text{Calibrate before use}
& 35.93  & 0.8   
& 77.30  & 1.5   
& 73.48  & 0.8   
& 83.32  & 0.5   
& 63.72  & 2.3   
& 80.54  & 0.8   
\\
\text{Logistic on logits}
& 36.86  & 0.8   
& 79.69  & 1.4   
& 73.75  & 0.7   
& 83.76  & 0.4   
& 64.47  & 2.3   
& 83.57  & 0.7   
\\
\text{Neural net on logits}
& 37.06  & 0.8   
& 78.59  & 1.4   
& 73.68  & 0.7   
& 83.67  & 0.4   
& 64.11  & 2.3   
& 83.08  & 0.7   
\\
\text{Logistic on last}
& 44.10  & 0.9   
& 71.20  & 1.7   
& 65.85  & 0.8   
& 81.48  & 0.5   
& 60.78  & 2.4   
& 81.52  & 0.8   
\\
\text{Neural net on last}
& 43.49  & 0.8   
& 79.93  & 1.4   
& 73.72  & 0.7   
& 83.89  & 0.4   
& 65.00  & 2.4   
& 85.13  & 0.7   
\\
\text{Logistic on last 10}
& 38.52  & 0.8   
& 79.66  & 1.4   
& 73.89  & 0.8   
& 83.65  & 0.4   
& 65.04  & 2.3   
& 84.13  & 0.7   
\\
\text{Neural net on last 10}
& 45.54  & 0.8   
& 80.02  & 1.5   
& 73.86  & 0.7   
& 83.67  & 0.4   
& 64.02  & 2.4   
& 84.91  & 0.7   
\\
\text{InnerThoughts}
& {\bfseries *47.24}  & 0.8   
& {\bfseries *80.27}  & 1.4   
& \bf 74.09  & 0.7   
& {\bfseries *83.99}  & 0.4   
& {\bfseries *66.57}  & 2.4   
& {\bfseries *87.24}  & 0.7   
\\\midrule
\text{Fine-tuning (QLoRA)}
& 48.50  & 0.8   
& 90.74  & 1.1   
& 75.00  & 0.7   
& 89.83  & 0.4   
& 69.30  & 2.3   
& 95.45  & 0.4   
\\\bottomrule
\end{tabularx}
\end{table*}

\section{Experiments}

\subsection{Setup}

We report results on a collection of six large multiple-choice question answering datasets: \textsc{AQuA}\footnote{\label{fn:MIT}MIT License} \citep{ling2017program}, \textsc{CosmosQA}\footnote{\label{fn:CC}CC 4.0 License} \citep{huang2019cosmos}, \textsc{MMLU}\footref{fn:MIT} \citep{hendrycks2020measuring}, \textsc{MedMCQA}\footnote{\label{fn:Apache}Apache 2.0 License} \citep{pal2022medmcqa},  \textsc{LogiQA}\footref{fn:CC} \citep{liu2020logiqa} and \textsc{HellaSwag}\footref{fn:MIT} \citep{zellers2019hellaswag}. These benchmarks were chosen for the range of tasks they cover, spanning mathematics, factual knowledge and reading comprehension, as well as the large size of their training sets. 

However, some of these datasets come with very small validation and test sets. Since the training sets are so large, in these situations we found it useful to randomly split the training set to remove unnecessary noise in method evaluation. Thus, we use a random 70\%-15\%-15\% train-validation-test split of the training set for \textsc{AQuA} and \textsc{MedMCQA}, while we use a 80\%-20\% train-validation split of the training set for \textsc{CosmosQA} and \textsc{HellaSwag}. The resulting split sizes are summarized in Table \ref{tab:dataset-sizes}. We note that \textsc{AQuA} additionally comes with rationales (``chains-of-thought''), which we do not use in our experiments. In every dataset the questions have four choices, except \textsc{AQuA}, in which the questions have five.

We use Llama3 70B \citep{dubey2024llama} as the large language model to extract the hidden states using the \texttt{Transformers} library, and implement predictor networks using the architecture detailed in Section \ref{sec:methodology} with $n_1=32$, $n_2=8$, LayerNorm as normalization layers and ReLU as activation functions. This was chosen by a small-scale hyperparameter search, although results did not particularly differ between the different choices. The predictor networks are trained using \texttt{Pytorch} for 50 epochs, with a learning rate of 1e-5 and early stopping using the validation set. The fine tuning results were obtained by the \texttt{SFTTrainer} utility from the \texttt{TRL} library, using QLoRA \citep{dettmers2024qlora} with adapter rank $r=1$. We train for three epochs\footnote{The computational requirements were extremely large even for 3 epochs, making training with more epochs unreasonably time-consuming.} with default hyperparameters and a batch size of 32. All experiments are performed on a machine with eighteen Intel Xeon 6140 CPUs and four Nvidia V100 32GB GPUs.

\subsection{Baselines and Variants}

Our approach consists of a neural network predictor taking as input the hidden states at the last token position of the prompt. We compare its performance against the following baselines.
\begin{itemize}[leftmargin=*,topsep=0pt,itemsep=1pt]
\item \textbf{Direct querying} The standard approach, where we softmax the logits without any transformation.
\item \textbf{Calibrate before use \citep{zhao2021calibrate}} Like in the paper, we use dummy prompts with [NA], [MASK] and an empty string for passages, questions and choices, and we average the probabilities of the three prompts as normalization factors $p_\text{norm}$. The final probabilities are computed as $\text{softmax}(p/p_\text{norm})$.

\item \textbf{Linear probe calibration \citep{abbas2024enhancing}} A logistic regression model on the logits.
\end{itemize}

We also compare with multiple \textbf{variants of the proposed method}. We test both neural networks and logistic regression models on (1) the final logits, (2) the last hidden state, and (3) the last 10 states. For the logistic regression on the final 10 states, we first perform a PCA on the states, as otherwise the number of features makes the fitting intractable. For the neural network on the last 8192 hidden state, we use a simple two layer feedforward network with 32 hidden neurons and a ReLU activation, while for the neural network on the last 10 states, which takes 10x8192-dimensional inputs, we use indeed the same architecture as in Section \ref{sec:methodology}. We are unable to provide results for a logistic regression model on all states, as in this situation, even the PCA approach is computationally intractable.

\begin{table*}[t]
\caption{Comparison of computational cost and gain for our method against QLoRA fine-tuning. We report (1) the total computational cost of training the model, in hours (using 18 Intel Xeon 6140 CPUs and 4 Nvidia V100 32GB GPUs -- see Appendix \ref{sec:timings} for costs for every method); (2) the gain of accuracy with respect to the Direct baseline; and (3) the resulting $\text{Cost}_{\text{InnerThoughts}}/\text{Cost}_{\text{FineTuning}}$ and $\text{Gain}_{\text{InnerThoughts}}/\text{Gain}_{\text{FineTuning}}$ ratios for every benchmark.}
\label{tab:gains}
\centering
\begin{tabularx}{\textwidth}{
    p{0.12\textwidth}
    >{\centering\arraybackslash}p{0.125\textwidth}
    >{\centering\arraybackslash}p{0.125\textwidth}|
    >{\centering\arraybackslash}p{0.125\textwidth}
    >{\centering\arraybackslash}p{0.125\textwidth}|
    >{\centering\arraybackslash}p{0.11\textwidth}
    >{\centering\arraybackslash}p{0.1\textwidth}
    }
\toprule
& \multicolumn{2}{c|}{Computational cost}
& \multicolumn{2}{c|}{Gain in accuracy}
& Comp.\ cost
& Acc.\ gain
\\
& InnerThoughts
& Fine-tuning
& InnerThoughts
& Fine-tuning
& ratio
& ratio
\\\midrule
\textsc{AQuA}
& \hphantom{0}35.4h
& \hphantom{0}171.3h
& +10.75
& +12.02
& 0.21
& 0.89
\\
\textsc{CosmosQA}
& \hphantom{0}14.5h
& \hphantom{00}85.2h
& +\hphantom{0}1.84
& +12.36
& 0.17
& 0.15
\\
\textsc{MMLU}
& \hphantom{0}70.7h
& 1097.6h
& +\hphantom{0}0.20
& +\hphantom{0}1.16
& 0.06
& 0.17
\\
\textsc{MedMCQA}
& \hphantom{0}57.9h
& \hphantom{0}223.2h
& +\hphantom{0}0.39
& +\hphantom{0}6.25
& 0.26
& 0.06
\\
\textsc{LogiQA}
& \hphantom{00}8.8h
& \hphantom{00}76.9h
& +\hphantom{0}2.42
& +\hphantom{0}5.15
& 0.12
& 0.47
\\
\textsc{HellaSwag}
& \hphantom{0}22.9h
& \hphantom{0}191.3h
& +\hphantom{0}5.04
& +13.24
& 0.12
& 0.38
\\\midrule
\textsc{Average}
& \hphantom{0}35.0h
& \hphantom{0}307.6h
& +\hphantom{0}3.44
& +\hphantom{0}8.36
& 0.16 
& 0.35
\\\bottomrule
\end{tabularx}
\end{table*}

\subsection{Results}\label{sec:results}

We report the accuracies of the different methods in Table \ref{tab:results-raw}. 
As can be seen, our InnerThoughts approach outperforms the previous baselines in the literature, Calibrate before use \citep{zhao2021calibrate} and Logistic on logits (a.k.a.\ Linear Probe Calibration) \citep{abbas2024enhancing}, sometimes quite significantly. The improvement in accuracy over the direct approach is particularly striking for \textsc{AQuA}, yielding over 10\% gain.

Since the method is trained on a collection of examples, it is valuable to compare the achieved performance against parameter-efficient fine-tuning methods, such as QLoRA \citep{dettmers2024qlora}. As discussed in Section \ref{sec:methodology}, although the method cannot be expected to achieve as high performance as fine-tuning methods, it should also be considerably cheaper computationally. A comparison of training costs between the two methods is provided as Table~\ref{tab:gains}. From this perspective, the comparison is favorable for our proposed approach: on average across datasets, InnerThoughts can obtain roughly 35\% of the gains of QLoRA at 16\% of the computational costs, a roughly 2x higher efficiency. Moreover, on one dataset (\textsc{AQuA}), we obtain almost as much improvement as supervised fine-tuning, at a fifth of the computational cost, a remarkable achievement. The approach can thus be seen as a cheaper alternative to adapter-based fine-tuning methods like LoRA/QLoRA, more accessible to the average hardware. This is particularly clear on the \textsc{MMLU} benchmark, where QLoRA takes over a thousand hours, which is prohibitive, whereas our approach only takes seventy, a much more realistic number.

\section{Analysis}

We now perform a detailed analysis of our method and results, aiming to answer two main questions: whether there is a pattern in which datasets yield the biggest gains, and why the inner states appear useful for prediction.

\subsection{Which datasets are most likely to benefit from our approach?}

It is clear, from Table \ref{tab:results-raw}, 
that the improvements vary widely between the different datasets. Is it possible to predict ahead of time which datasets are most likely to benefit from our approach?

\begin{figure}[t]
    \includegraphics[width=\columnwidth]{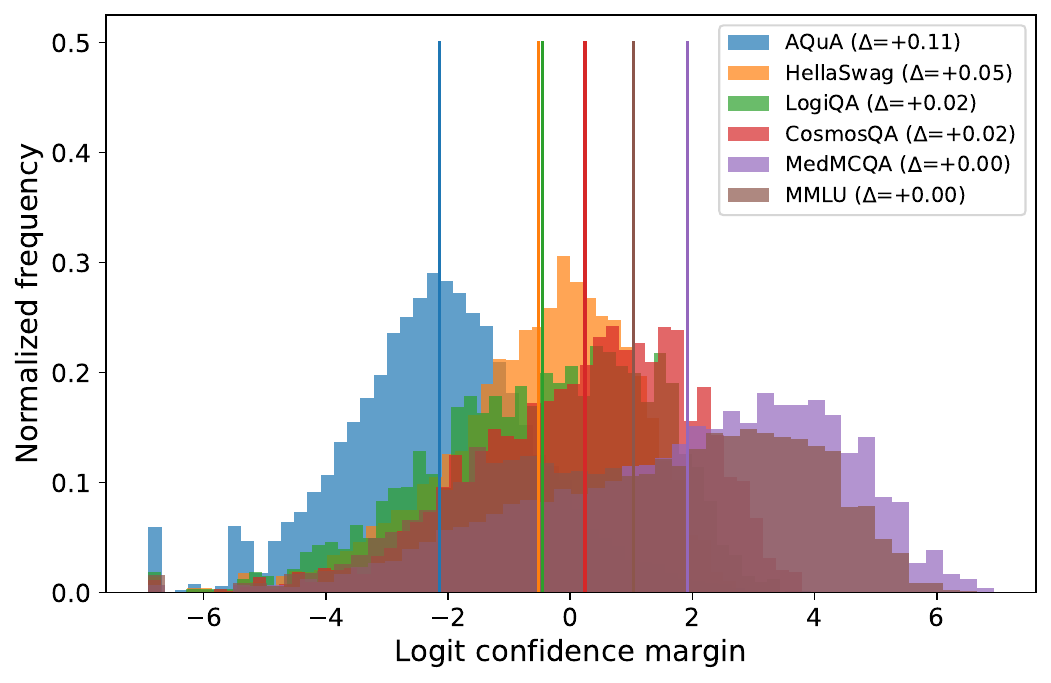}
    \caption{Histograms of logit-transformed confidence margins (difference between top two probabilities) among the answer distributions outputted by the standard LLM approach (``Direct method'') on the various test datasets. We also provide, in the legend, the gain in accuracy ($\Delta$) associated with applying InnerThoughts to each dataset. The vertical bars represent the mean of the logit-margin distributions.}
    \label{fig:logits-margin}
\end{figure}

One intriguing possibility is that the benefits of our approach are related to the degree of confidence the original LLM had in its answers. To see this, we compute for any test question the confidence margin of the LLM, that is, the difference between the top and second highest probabilities $m=p_{(1)}-p_{(2)}$ in the distribution $p$ outputted by the LLM using the standard strategy (``Direct''). Since these margins $m$ are limited between 0 and 1, which tends to crowd the plots, we find it useful to plot instead the logit transform of the margin $\text{logit}(m)=\log(m/(1-m))$, so as to more clearly see patterns between datasets. The resulting plot is provided as Figure \ref{fig:logits-margin}. 

As can be seen, the gains seem roughly correlated with the confidence margin of the LLM: the datasets where InnerThoughts bring the greatest gains are precisely those for which the LLM was hesitating the most between the top and second highest answer (small confidence margin). In particular, the ordering between the datasets in Figure \ref{fig:logits-margin} is almost the same as the ordering between datasets according to gains with InnerThoughts, with only \textsc{MedMCQA} and \textsc{MMLU} reversed, which are almost zero in any case. This suggests that InnerThoughts is useful precisely for the datasets where the LLM hesitates between its answers.

 \begin{figure*}[t]
    \includegraphics[width=\textwidth]{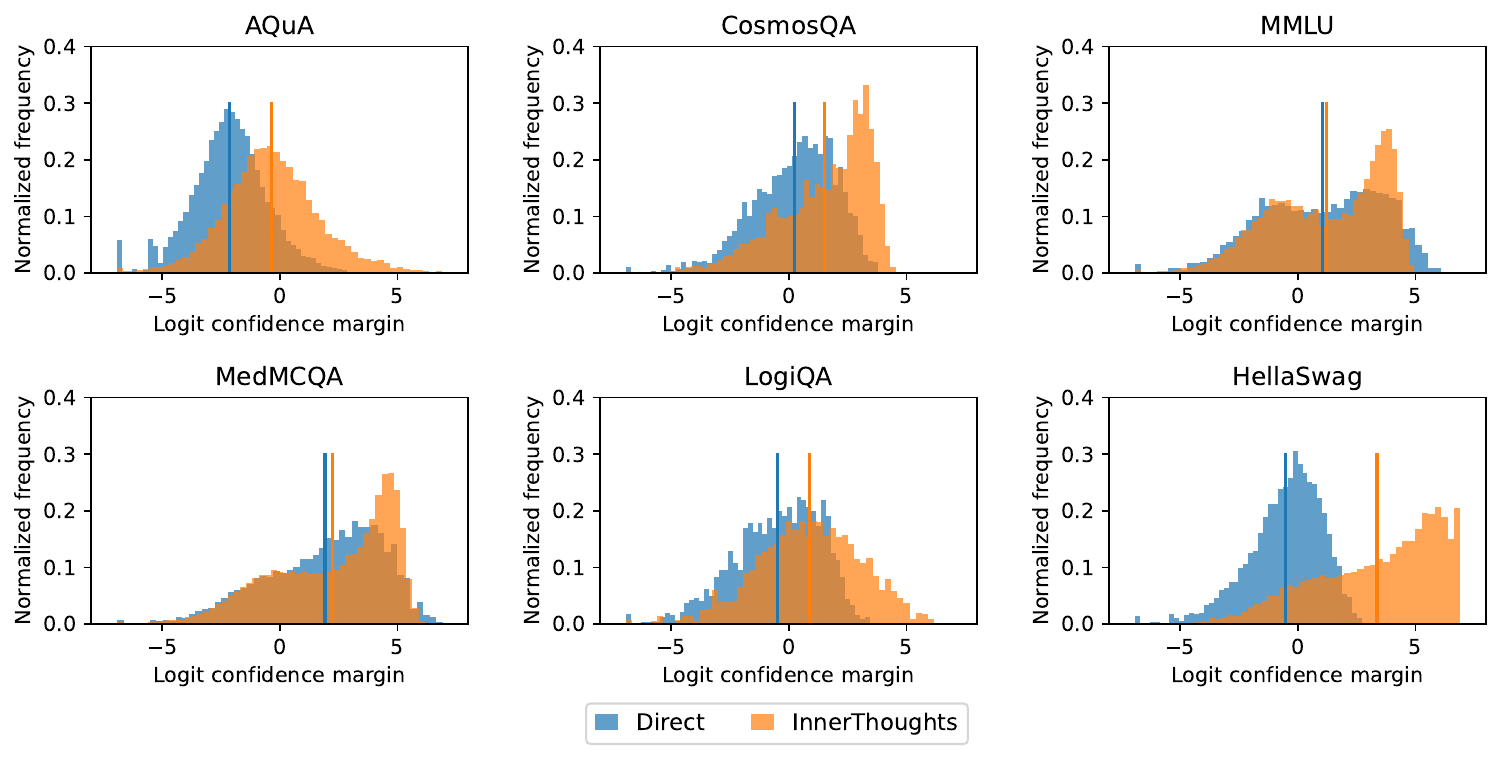}
    \caption{Histograms of the logit-transformed confidence margins, $\text{logit}(p_{(1)}-p_{(2)})$, on the answer distributions $p$ produced by the standard strategy (``Direct'') and InnerThoughts. The vertical bars represent the mean of the logit-margins for each distribution.}
    \label{fig:logits-margin-BA}
\end{figure*}

It is also enlightening to compare what happens to those confidence margins after applying InnerThoughts, on each benchmark, which is summarized in Figure \ref{fig:logits-margin-BA}. As can be seen, the degree of confidence in the answer (represented by the logit-margin distributions) is significantly increased precisely on those datasets where InnerThoughts bring the largest gains. It appears that InnerThoughts not only corrects the answers but also increases the confidence of the LLM on the benchmarks where it has an effect.

These findings in turn beg the question as to why the LLM would have systematically lower confidence margins on some benchmarks rather than some others. Analyzing some examples, we can provide some hypotheses.

\begin{table}[t]
\caption{An example from the HellaSwag dataset.}
\label{tab:hellaswag-example}
\vspace{-5pt}
\begin{tcolorbox}[colback=blue!5!white]
\textbf{Passage}: A cartoon animation video is shown with people wandering around and rockets being shot. \\ \textbf{Question}: Which choice best continues the passage? \\ A. fight robots of evil and ends with a to be continued.\\ B. are then shown in closeups shooting a shot put.\\ C. push a child in a speedboat in the water.\\ D. look in the cameraman's eye and smile.\\ \textbf{Answer: }A
\end{tcolorbox}
\vspace{-10pt}
\end{table}

First, it is striking that the biggest gains are realized on the only dataset (\textsc{AQuA}) whose answers are numerical, since the questions are essentially elementary algebra questions. Considering that those questions are fundamentally mathematical rather than linguistic, it is perhaps unsurprising that the LLM hesitates a lot about the answers. In addition, the second best performing dataset, \textsc{HellaSwag}, is a dataset of problems designed to test passage-completion, where the model is provided a scenario and a list of potential continuations. However, reading the problems, we see that the continuations are often semantically and grammatically disconnected from the given context. 

For example, consider the example in Table~\ref{tab:hellaswag-example}.
Even for a human, answering the question is difficult, as none one of the options are grammatically consistent with the initial passage, and each introduce new semantic contexts (e.g. robots, shot-put and speedboat). In fact, from a purely linguistic basis, it would be reasonable to expect the answer to be D, since it is the phrase least in need of a subject, and also the most lexically related (``cameraman'' is closely related to ``video''). To understand the correct answer to be A requires some amount of mental visualization beyond mere linguistic considerations.

Thus in some sense, both these datasets require strong non-linguistic skills, namely algebraic (\textsc{AQuA}) and visualization (\textsc{HellaSwag}) skills. In turn, these missing, or weaker, skills lead the LLM to have lower confidence margins in its answers. InnerThoughts, which can be regarded as an adapter built on top of the LLM, perhaps helps precisely by learning the skills necessary to solve these problems more successfully.

\subsection{Why use all the layers' hidden states?}

Another question that stands out is that, since the final layer states are themselves computed from the lower layer states, why does using these extra inner states bring any additional improvement? In particular, as transformer layers themselves involve residual connections, one could imagine that all information from earlier layers is preserved in the final hidden state.

A first reason why performance might still improve by providing the extra inner states to the model is that even if information is preserved, it might be clustered in a way that is useful for language modeling but inconvenient for training a question-answering predictor module.
Thus, even in this scenario, it might be worthwhile to provide the whole sequence of additive contributions rather than their final ``mashup''.

A second, and more critical reason is that there might simply not be enough dimensions to conserve all the information from layer to layer, despite the residual connection. For example, in Llama3 the model has 80 transformer layers with 64 attention heads each, and each attention head provides a 128-dimensional contribution projected down into the 8192-dimensional hidden state space. For no contribution to overlap, one would need 80x64x128=655,360 different dimensions, much larger than the 8192 dimensions in the hidden space. Thus, it appears unavoidable that at least some information in the hidden state gets overwritten from layer to layer by deeper attention heads.

Finally, a third reason is that as several prior works such as \citet{chen2023bigger} and \citet{todd2024function} have remarked, some layers in LLMs appear systematically associated with certain abilities more than others. If true, if would be reasonable that their insights would be best used by extracting them directly, rather than by their collective contribution in one hidden-state at the final layer of the model.

\begin{figure}
    \includegraphics[clip, trim=7pt 0pt 20pt 0pt, width=\columnwidth]{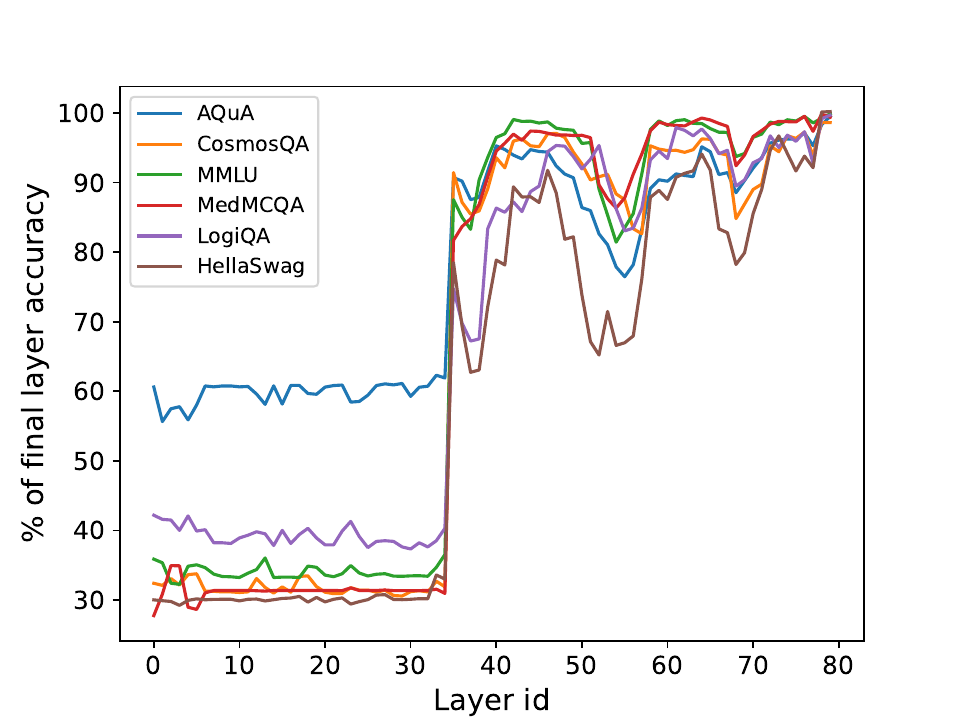}
    \caption{Final accuracy when using the hidden-state at each layer of the LLM individually to compute logits using the tokenization matrix and make the classification, as a percentage of full direct-prompting accuracy for each dataset.}
    \label{fig:layer_acc}
\end{figure}

To test this latter hypothesis, we investigated the effect of each layer on the final prediction using the Logit Lens \citep{logitlens, belrose2023eliciting} method, by extracting a distribution over answer tokens by using an earlier hidden state rather than the one of the last layer. The accuracy of each layer on the test benchmarks is provided as Figure \ref{fig:layer_acc}. As can be seen, on each benchmark the predictive performance of the inner layers is marginal, until it hits precisely layer 35, where it jumps to a near-final accuracy. This suggests that the layers, especially layer 35 in the Llama3 architecture, have differing value in solving our question-answering tasks.

In addition, we performed two supplemental analyses in Section \ref{sec:inner-analysis} of the appendix: one to quantify whether InnerThoughts is effectively able to disentangle the additive contributions from each transformer layer, and another to identify which layers contribute the most to the predictions. The results show that the model appear to be already disentangling the representations well, and using all the inner hidden states, especially those after layer 35, in agreement with Figure \ref{fig:layer_acc}.

\section{Limitations and Extensions}\label{sec:limitations}

One limitation of the current work is that we restricted ourselves to multiple-choice question answering problems. However, by changing the predictor module from a classification to a regression model, one could imagine extending the work to tasks whose answer is an arbitrary number, like the arithmetic problems of GSM8k \citep{cobbe2021gsm8k}. In addition, our results are limited to a ``zero-shot'' setting, but one could imagine extending the work to a few-shot setting where prior question-answer example pairs are provided at test time, again without much difficulty.

Another extension we did not consider in our work is how to integrate our method with the popular chain-of-thought framework \citep{wei2022chain}. One possibility would be to apply the predictor module after the chain-of-thought has been generated, although it is not clear that there are as many biases to be corrected in this scenario, hence as much to gain.

Finally, although throughout this work we considered supervised fine-tuning as a competing approach, we could have imagined instead applying InnerThoughts to the fine-tuned model itself. In this situation, it is possible that additional gains might be obtained on top of those brought by fine-tuning. In fact, we could even imagine fine-tuning the LLM and training the predictor module simultaneously, in an end-to-end manner.

\section{Conclusion}

In this work, we develop a novel strategy for improving large language model accuracy when answering multiple-choice questions. Using a set of labeled question-answer pairs, a small predictor neural network is trained to directly predict the answer from the collection of hidden states generated by every transformer layer at the last token position. On a collection of benchmarks, the approach is shown to provide substantial improvements over the standard approach, achieving sometimes similar performance to adapter-based fine-tuning methods, at a fraction of the computational costs.

\bibliography{references}
\bibliographystyle{apalike}

\newpage

\appendix
\onecolumn
\section{Example prompts}\label{sec:example-prompts}

We present examples of prompts for each dataset used in this paper, which follow the guidelines of \citet{robinson2023leveraging}.

\begin{table}[ht]
\centering
\caption{Multiple-choice question example from the AQuA benchmark. The correct answer is D.}
\label{tab:aqua-example}
\vspace{-5pt}
\ttfamily
\frenchspacing 
\begin{tabularx}{\linewidth}{>{\raggedright\arraybackslash}X}
\toprule
Question: Two persons start running simultaneously around a circular track of length 380 m from the same point at speeds of 24 kmph and 28 kmph. When will they meet for the first time any where on the track if they
are moving in the opposite direction ?\\
A. 144\\
B. 36\\
C. 124\\
D. 26\\
E. 38\\
Answer:
\\\bottomrule
\end{tabularx}
\end{table}

\begin{table}[ht]
\centering
\caption{Multiple-choice question example from the CosmosQA benchmark. The correct answer is D.}
\label{tab:cosmosqa-example}
\vspace{-5pt}
\ttfamily
\frenchspacing 
\begin{tabularx}{\linewidth}{>{\raggedright\arraybackslash}X}
\toprule
Passage: I listened to dry your eyes mate by the streets on the way home . And I thought I was doing a good job of getting over it . Then I came back and a whole new pain was brought into the situation .
Question: What will I want to do next ?\\
A. None of the above choices .\\
B. Make a hair appointment .\\
C. Talk to someone about a new job .\\
D. Talk to someone about my feelings .\\
Answer:
\\\bottomrule
\end{tabularx}
\end{table}

\begin{table}[ht]
\centering
\caption{Multiple-choice question example from the MMLU benchmark. The correct answer is B.}
\label{tab:mmlu-example}
\vspace{-5pt}
\ttfamily
\frenchspacing 
\begin{tabularx}{\linewidth}{>{\raggedright\arraybackslash}X}
\toprule
Question: From what does a chain of command extend?\\
A. Bottom to top\\
B. Top to bottom\\
C. Diagonally\\
D. Laterally\\
Answer:
\\\bottomrule
\end{tabularx}
\end{table}

\begin{table}[ht]
\centering
\caption{Multiple-choice question example from the MedMCQA benchmark. The correct answer is B.}
\label{tab:medmcqa-example}
\vspace{-5pt}
\ttfamily
\frenchspacing 
\begin{tabularx}{\linewidth}{>{\raggedright\arraybackslash}X}
\toprule
Question: The mandibular molars generally have\\
A. Two roots \& two canals \\
B. Two roots \& three canals\\
C. Three toots \& two canals\\
D. None of the above\\
Answer:
\\\bottomrule
\end{tabularx}
\end{table}

\begin{table}[ht]
\centering
\caption{Multiple-choice question example from the LogiQA benchmark. The correct answer is C.}
\label{tab:logiqa-example}
\vspace{-5pt}
\ttfamily
\frenchspacing 
\begin{tabularx}{\linewidth}{>{\raggedright\arraybackslash}X}
\toprule
Passage: Studies suggest that, for the vast majority of people who have normal blood pressure, any amount of sodium greater than that required by the body is simply excreted and does not significantly raise blood pressure. So only persons who have high blood pressure and whose bodies are incapable of safely processing excess sodium need to restrict their sodium intake.\\
Question: Which one of the following, if true, would most seriously weaken the argument?\\
A. Every human being has a physiological need for at least some sodium.\\
B. Any sodium not used by the body will increase blood pressure unless it is excreted.\\
C. Excess sodium intake over time often destroys the body's ability to process excess sodium.\\
D. High blood pressure is more harmful than was previously believed.\\
Answer:
\\\bottomrule
\end{tabularx}
\end{table}

\begin{table}[ht!]
\centering
\caption{Multiple-choice question example from the HellaSwag benchmark. The correct answer is B.}
\label{tab:hellaswag-example2}
\vspace{-5pt}
\ttfamily
\frenchspacing 
\begin{tabularx}{\linewidth}{>{\raggedright\arraybackslash}X}
\toprule
Passage: [header] How to apply foundation [title] Wash your face. [step] Cleaning your skin properly will remove dirt and oil as well as any previous makeup. Make sure to choose a product designed for your skin type : [substeps] Use cleansing water to reduce redness, as this suds-free cleanser is infused with anti-inflammatory agents that will calm your skin.
Question: Which choice best continues the passage?\\
A. Gently pat your face dry with a soft soft sponge or microfiber cloth and do not wash your face with oil. [title] Blend the foundation remover with an exfoliating cleanser.\\
B. Cleansing balms, packed with emollients, are great for adding moisture to dry skin. Pick a cleansing mud for oily skin, as the charcoal and clay will remove excess oil from your pores.\\
C. [title] Apply a single layer of foundation all over your face. [step] This is a one-time step that you will want to take as you will be applying foundation to two different areas.\\
D. [substeps] Use facial cleansers that are set at a ph level between 1 and 3 or 5 which help to replenish the skin's nutrients and help to prepare your skin for foundation. [title] Stretch your skin with your fingertips to make it firmer.\\
Answer:
\\\bottomrule
\end{tabularx}
\end{table}

\newpage
~\newpage

\section{Computational cost of the methods}\label{sec:timings}

We summarize in Table \ref{tab:timings} the computational cost of training each method, in hours.

\begin{table*}[ht]
\caption{Computational cost of training each model, in hours (using 18 Intel Xeon 6140 CPUs and 4 Nvidia V100 32GB GPUs). On the predictor-based models, the vast majority of the computational cost (98\%) is spent extracting the hidden states for the training set. No training is required for Direct and Calibrate Before Use.}
\vspace{1mm}
\label{tab:timings}
\centering
\aboverulesep = 0.5mm
\belowrulesep = 1mm
\begin{tabularx}{\textwidth}{
    p{0.2\textwidth}
    >{\centering\arraybackslash}p{0.1\textwidth}%
    >{\centering\arraybackslash}p{0.11\textwidth}%
    >{\centering\arraybackslash}p{0.1\textwidth}%
    >{\centering\arraybackslash}p{0.11\textwidth}%
    >{\centering\arraybackslash}p{0.1\textwidth}%
    >{\centering\arraybackslash}p{0.11\textwidth}%
    }
\toprule
& \textsc{AQuA}
& \textsc{CosmosQA}
& \textsc{MMLU}
& \textsc{MedMCQA}
& \textsc{LogiQA}
& \textsc{HellaSwag}
\\
\midrule
\text{Direct}
& -
& -
& -
& -
& -
& -
\\
\text{Calibrate before use}
& -
& -
& -
& -
& -
& -
\\
\text{Logistic on logits}
& 34.6h
& 14.3h
& 69.9h
& 56.3h
&  8.7h
& 22.4h
\\
\text{Neural net on logits}
& 34.7h
& 14.3h
& 70.1h
& 56.6h
&  8.8h
& 22.6h
\\
\text{Logistic on last}
& 36.2h
& 14.3h
& 72.5h
& 60.0h
&  8.8h
& 22.7h
\\
\text{Neural net on last}
& 34.7h
& 14.3h
& 70.0h
& 56.5h
&  8.8h
& 22.6h
\\
\text{Logistic on last 10}
& 44.9h
& 14.4h
& 70.4h
& 58.0h
&  8.8h
& 22.5h
\\
\text{Neural net on last 10}
& 34.8h
& 14.3h
& 70.1h
& 56.6h
&  8.8h
& 22.6h
\\
\text{InnerThoughts}
& 35.4h
& 14.5h
& 70.7h
& 57.9h
&  8.8h
& 22.9h
\\\midrule
\text{Fine-tuning (QLoRA)}
&  171.3h
&   85.2h
& 1097.6h
&  223.2h
&   76.9h
&  191.3h
\\\bottomrule
\end{tabularx}
\end{table*}

\section{Statistical calibration}\label{sec:statistical-calibration}

Our work ties in with related literature on large language model calibration, which we can define broadly as mitigation of bias coming from training for autoregressive language modeling but testing for the true objective of question-answering. 
We note that despite the name, this concept is unrelated to statistical calibration, which is the property of a classifier of having the probabilities of the outcomes match the probabilities predicted by the model. This is typically evaluated with a quantitative measure such as the Brier score.

As a curiosity, we investigated whether our approach can also improve statistical calibration. Brier scores per datasets are summarized in Table \ref{tab:brier}. As can be seen, in general the method does not appear to change the amount of statistical calibration in a noticeable pattern, with only MMLU and LogiQA offering any big change, in opposite directions.

\begin{table*}[ht]
\caption{Brier scores for the competing LLM calibration methods evaluated in the paper, on the different datasets. Lower is better.}
\label{tab:brier}
\centering
\aboverulesep = 0.5mm
\belowrulesep = 1mm
\begin{tabular}{lcccccc}
\toprule
& \textsc{AQuA}
& \textsc{CosmosQA}
& \textsc{MMLU}
& \textsc{MedMCQA}
& \textsc{LogiQA}
& \textsc{HellaSwag} 
\\ \midrule
Direct        & 0.138     & 0.192     & 0.165     & 0.097     & 0.216     & \bf 0.133 \\
Calibrate     & 0.138     & 0.195     & 0.166     & 0.098     & 0.215     & 0.135     \\
Lin Probe     & 0.124     & \bf 0.188 & 0.106     & 0.097     & \bf 0.213 & 0.137     \\
InnerThoughts & \bf 0.124 & 0.190     & \bf 0.080 & \bf 0.097 & 0.239     & 0.140     \\ \bottomrule
\end{tabular}
\end{table*}

\section{How do the predictor networks use the inner representations?}\label{sec:inner-analysis}

In this section we attempt to provide further insight on how the predictor networks make use of the extra information in the earlier layers for performance gain, with two analyses. As before, all analyses are performed on a Llama3 70B model.

The first analysis aim to establish whether the model is already able to disentangle the additive contributions from the transformer layers from the hidden states themselves, which are their cumulative sum. To show this, we trained a model directly on the additive contributions, rather than the hidden states themselves. The results of this "DiffHiddenThoughts" method is shown in Table \ref{tab:diffhiddenthoughts}, with the LLM baseline shown as a reference.
As can be seen, the results are better on one dataset (LogiQA) but equal or slightly worse on the others, and overall it does not seem to make much of a difference. Thus the existing model seem to already be able to disentangle the additive contributions well.

\begin{table}[ht]
\caption{Accuracy of a predictor network trained on the additive contributions from each transformer layer (``DiffHiddenThoughts'') with the regular model trained on the hidden thoughts (``HiddenThoughts'').}
\label{tab:diffhiddenthoughts}
\centering
\aboverulesep = 0.5mm
\belowrulesep = 1mm
\begin{tabular}{lcccccc}
\toprule
& \textsc{AQuA}
& \textsc{CosmosQA}
& \textsc{MMLU}
& \textsc{MedMCQA}
& \textsc{LogiQA}
& \textsc{HellaSwag} 
\\ \midrule
    Direct & 36.47 & 78.38 & 73.88 & 83.59 & 64.25 & 82.20 \\ 
    InnerThoughts & \bf 47.24 & \bf 80.27 & \bf 74.09 & \bf 83.99 & 66.57 & 87.24 \\ 
    DiffInnerThoughts & 46.98 & 79.73 & 73.99 & 83.84 & \bf 68.67 & \bf 87.34 \\ 
\bottomrule
\end{tabular}
\end{table}

\begin{figure}[ht]
    \includegraphics[clip, trim=5pt 5pt 10pt 10pt, width=0.7\textwidth]{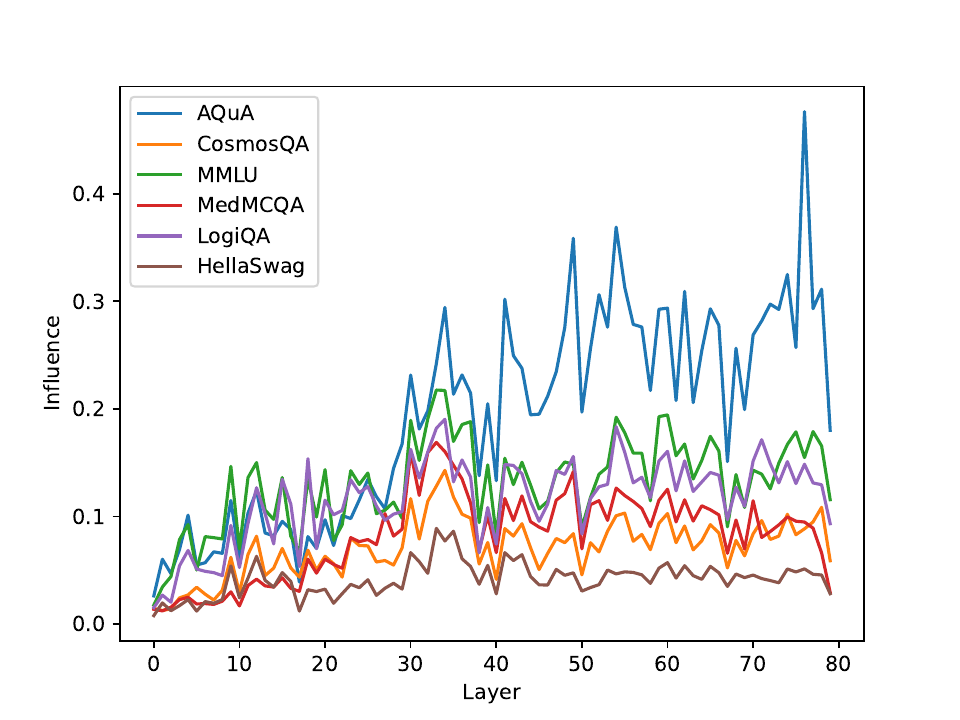}
    \centering
    \caption{Influence score $\sum_{k=1}^{8192}\big|{\partial f(h)}/{\partial h_{lk}}\big|^2$ for each layer on the predictions made by the InnerThoughts predictor module. Higher means more influence of the hidden state at that layer on the prediction.}
    \label{fig:grad-influence}
\end{figure}

The other analysis we performed aimed to investigate which hidden states were most influent in the predictions of the model, which we did by investigating the size of the gradients of the prediction with respect to the features. More precisely, we computed, for each layer,
\begin{align*}
    \text{Influence}_{l} = \sum_{k=1}^{8192}\Big|\frac{\partial f(h)}{\partial h_{lk}}\Big|^2
\end{align*}
where $f$ denote the predictor module and $h_{lk}$ denotes the hidden state at the $l^\text{th}$ layer and $k^\text{th}$ component at the last token position (as in the paper). The resulting influence metrics per layer are summarized in  Figure \ref{fig:grad-influence} for each dataset.
As can be seen, although the earliest layers tend to be less influent, in general the middle and later layers tend to be relatively equally important to the model. Thus, in particular, it doesn't appear that the model is only focusing on layer 80, but it seems to be using the earlier layer information just as well.

\section{Alternative architecture for the predictor network}

We also ran a small experiment for alternative architectures for the predictor module of InnerThoughts. In this experiment, the predictor module consists of a self-attention layer with a single head which takes the sequence of the 80 hidden states projected down to 32 dimensions as input. We added a learned "classification token" to each sequence, bringing it to a length of 81, and the final output at the last position was given to a single 32-dimensional fully-connected layer to perform classification. We added the standard sinusoidal positional embeddings to each input in the sequence. There is a layer-normalization layer before the self-attention block and a ReLU activation function prior to the prediction fully-connected layer. The learning rate used was 1e-5 which is the same as the model highlighted in the paper. The results are summarized in Table \ref{tab:self-attention}. As can be seen, the results are slightly worse but as we did not have time to thoroughly optimize this architecture, it is possible that with more effort performance could be brought on par with the current results (or better).

\begin{table}[ht]
\caption{Accuracy of the Self-Attention architecture compared to the architecture proposed in Section \ref{sec:methodology}.}
\label{tab:self-attention}
\centering
\aboverulesep = 0.5mm
\belowrulesep = 1mm
\begin{tabular}{lcccccc}
\toprule
& \textsc{AQuA}
& \textsc{CosmosQA}
& \textsc{MMLU}
& \textsc{MedMCQA}
& \textsc{LogiQA}
& \textsc{HellaSwag} 
\\ \midrule
        InnerThoughts & \bf 47.24 & \bf 80.27 & \bf 74.09 & \bf 83.99 & \bf 66.57 & \bf 87.24 \\ 
        Self-Attention & 45.79 & 79.40 & 74.01 & 83.79 & 66.22 & 85.73 \\ \bottomrule 
    \end{tabular}
\end{table}

\end{document}